\documentclass[runningheads]{llncs}
\usepackage[T1]{fontenc}
\usepackage[utf8]{inputenc}
\usepackage{graphicx}
\usepackage{hyperref}
\usepackage{color}
\usepackage{booktabs}
\usepackage{multirow}
\usepackage{amsmath}
\usepackage{amssymb}
\usepackage{enumitem}
\usepackage{xcolor}
\usepackage{microtype}
\usepackage{siunitx}

\begin{document}
\title{LLM-as-a-Discriminator: When Synthetic Tables Still Look Real}
\titlerunning{LLM-as-a-Discriminator}
%
\author{Manel Slokom\inst{1}\orcidID{0000-0002-9048-1906} \and
Malek Slokom\inst{2} \and
Thierno Kante\inst{3}}
\authorrunning{M. Slokom et al.}
%
\institute{Vrije Universiteit, Amsterdam \email{manel.slokom@live.fr} \and
Equativ, Paris \email{slokom.malek@livgmaile.com}
\\
\and
EDICIA, Nantes \email{thierno.kante@edicia.fr}}
%
\maketitle              
\begin{abstract}
Privacy and data sharing are often in tension. Many organizations use synthetic data to reduce privacy risk and still share useful data. 
For tabular data, auditing privacy remains hard. 
In many cases, even humans cannot easily tell if a table is real or synthetic.
In this paper, we propose a method based on LLM discrimination. 
We ask an LLM to classify each table sample as \texttt{REAL} or \texttt{SYNTHETIC}.
We test two settings: C1 with table only, and C2 with table plus distributional metadata. 
We use LLaMA as an open model and Gemini as a reference model.
In our experiments, we run three synthesis models, CTGAN, TVAE, and Gaussian Copula, on two public datasets, UCI Adult and ACS Census. 
We collect \textbf{451 valid trials}. Our results show clear differences between models. 
On Adult, LLaMA reaches DRS=0\% in reported cells, while Gemini reaches DRS=100\% for CTGAN and TVAE. 
On Census, LLaMA predicts \texttt{SYNTHETIC} for most samples, while Gemini stays high in C1 but drops for CTGAN and TVAE in C2.
We also compare with a classifier two-sample test (C2ST) and record linkage as distributional baselines, and with a human pilot of 2 annotators and 240 trials. 
Our results show that LLM discrimination is a practical privacy audit signal when model choice,
per provider reporting, and data encoding are handled with care.
For reproducibility, code and experiment scripts are available at
\url{https://github.com/SlokomManel/LLM-as-a-Discriminator}.
\end{abstract}
\section{Introduction}

Organizations increasingly turn to synthetic tabular data as a privacy-preserving alternative to real microdata. 
The key practical question is whether an informed adversary can still tell which records are real. 
If so, the protection offered may be weaker than assumed.

Classical Statistical Disclosure Control (SDC) tools, e.g., k-anonymity, microaggregation, record linkage~\cite{hundepool2012statistical,Torra2017,shlomo2022How}, were designed for masked or perturbed data without having generative models in mind. 
The rise of deep generative models has prompted a shift toward membership inference attacks (MIAs) as a privacy risk measure~\cite{stadler2020synthetic,shokri2015quantifying}. 
MIA is framed as a binary classification: given a target record and a synthetic release, did the record belong to the original training set? 
State-of-the-art shadow modeling approaches achieve non-trivial true positive rates even in the black-box setting, though no single attack dominates across generators and datasets~\cite{Shokri2017Membership}. 
Attribute inference~\cite{Bargav2022Are} and linkage attack further confirm that synthetic data faces privacy-utility trade-offs analogous to classical anonymization. 
Yet all these methods share a common limitation: they do not capture whether a \emph{human or AI observer} can directly distinguish a synthetic table from a real one. 

Large language models (LLMs) are increasingly used as automated evaluators.
In~\cite{zheng2023judging}, the authors showed that strong LLM judges match human preferences at over 80
However, using an LLM as a black-box \emph{discriminator} for privacy auditing of synthetic tabular releases has not been studied systematically.
We address this gap with an \emph{LLM-as-Discriminator} protocol under two threat conditions: table only (C1) and table plus distributional metadata (C2).
This framing mirrors a realistic adversary with access to the release but not to generator internals, and yields interpretable outputs (verdict, confidence) that practitioners can inspect.
We make four contributions:
(i)~an \emph{LLM-as-Discriminator\footnote{Code available at
\url{https://github.com/SlokomManel/LLM-as-a-Discriminator}.}} protocol for black-box privacy
auditing under two threat conditions: table only (C1) and table plus
metadata (C2);
(ii)~an evaluation on \textbf{451 valid trials} across two datasets,
three synthesis methods, and two LLM families, showing strong model-
and encoding-dependence;
(iii)~a balanced \emph{Disclosure Risk Score} (DRS) compared with
empirical baselines (C2ST and record linkage), with directional
alignment reported;
(iv)~a controlled human pilot (240 trials) showing that LLaMA lags
human annotators while Gemini matches or exceeds human performance.
 
Our experimental evaluation is organized around five research questions,
detailed in Section~\ref{sec:results}.
The remainder of this paper is organized as follows.
Section~\ref{sec:related} reviews related work on disclosure risk, data
synthesis, and LLM-based evaluation.
Section~\ref{sec:method} describes the LLM-as-Discriminator protocol
and the Disclosure Risk Score.
Section~\ref{sec:setup} presents the experimental setup.
Section~\ref{sec:results} reports our results.
Section~\ref{sec:discussion} discusses our findings and connects them
to classical privacy measures.
Section~\ref{sec:conclusion} concludes and outlines directions for
future work.
Additional analysis is provided in the Appendix~\ref{sec:appendix}.



\section{Background and Related Work}
\label{sec:related}

In this section, we review related work in three parts: (1) privacy risk measures, (2) synthesis methods, and (3) LLM-as-evaluator. 

\subsection{Privacy Risk Measures for Statistical Databases}
Privacy risks take two main forms~\cite{Torra2017Risk}: identity
disclosure, where an adversary links external information to a released record, and attribute disclosure, where an adversary infers sensitive information about an individual.
 
\textit{SDC-based measures.}
$k$-anonymity~\cite{sweeney2002achieving}, $\ell$-diversity, and
$t$-closeness~\cite{Torra2017Risk} measure identity and attribute
disclosure risk through indistinguishability constraints.
CAP~\cite{Mark2014CAP,Little2022TCAP} estimates the probability that an adversary correctly guesses a sensitive attribute value.
These measures are well-established but were designed for masked or
perturbed data, not generative models.
 
\textit{Attack-based measures.}
MIAs~\cite{Shokri2017Membership} infer whether a target record
belonged to the generator's training set.
Attribute inference attacks~\cite{Bargav2022Are,Salamatian2013elephant}
target sensitive attribute values rather than membership.
Record linkage~\cite{garfinkel2015identification,powar2023sok} measures the risk of matching a synthetic record to an external database.
We use MIAs and record linkage as formal baselines.
 
To the best of our knowledge, none of these methods captures whether an informed observer can directly distinguish a synthetic table from a real one after release.
We address this gap.
\subsection{Synthetic Tabular Data Generation}
\label{sec:related:synthesis}
Synthesis methods for tabular data differ in how they model and sample from the underlying data distribution.
We provide a brief overview of the main approaches.

\textit{Statistical methods.}
In~\cite{patki2016synthetic}, the authors use Gaussian copulas and
parametric marginal models to reproduce pairwise column correlations.
CART-based synthesizers draw records from estimated conditional
distributions via recursive partitioning~\cite{reiter2005using,unitedsynthetic}.
Both approaches are transparent, but their outputs carry predictable statistical structure that attack-based methods can exploit.

\textit{GAN-based methods.}
CTGAN~\cite{NIPS2019ctgan} adapts the conditional GAN paradigm to
heterogeneous tabular data using mode-specific normalization of
continuous columns and conditional vector training.
TVAE applies a variational autoencoder to learn a latent embedding of the joint distribution.
Both methods capture complex inter-column dependencies more faithfully than statistical baselines.

\textit{Diffusion and language model methods.}
TabDDPM~\cite{kotelnikov2023tabddpm} applies denoising diffusion
probabilistic models to tabular data and achieves strong fidelity on standard benchmarks.
GReaT~\cite{borisov2023GReat} serializes tabular rows as natural
language sentences and fine-tunes a large language model to generate new rows by token-level completion.
Records produced by these methods can be semantically plausible and
resist purely distributional detection, raising questions that statistical evaluation protocols were not designed to answer.
\subsection{LLMs as evaluators}
\label{sec:related:llm}
LLMs have shown strong performance across a wide range of tasks and
have been extended to tabular data processing~\cite{zhao2026survey,fanglarge}.
More recently, they have been used as automated evaluators replacing or complementing human annotation~\cite{Hu2025llmjudge}.
Zheng et al.~\cite{zheng2023judging} showed that GPT-4 judgments agree with human preferences at over 80\%, establishing LLM-as-a-Judge as a scalable alternative to human annotation.
Chiang and Lee~\cite{chiang2023can} found high LLM-human agreement on well-defined binary tasks and documented systematic biases such as position bias and verbosity preference.
We follow this binary-task framing and apply it to a new setting:
discriminating real from synthetic tabular records.
 
 
\textit{Interpretability of LLM judgments.}
LLM judges produce a verdict alongside a natural language rationale that explains their decision~\cite{han2025judge}. 
This property is useful in privacy auditing, where practitioners need not only a discrimination signal but also an explanation of why a record appears synthetic.
Practitioners need not only a discrimination signal but also an explanation of why a record appears synthetic.
Our proposed method elicits a verdict, a confidence score, and a rationale in a single prompt.
We use these rationales to analyze model reasoning across threat conditions.

\section{LLM-as-Discriminator Framework}
\label{sec:method}

We frame privacy auditing as a binary classification task.
Given a tabular sample $T$ drawn from either a real dataset
$\mathcal{R}$ or a synthetic dataset $\mathcal{S}$, a discriminator
$\mathcal{D}$ must output three things: a verdict
$v \in \{\texttt{REAL}, \texttt{SYNTHETIC}\}$, a confidence score
$c \in [0, 100]$, and a natural-language rationale.
We instantiate $\mathcal{D}$ with large language models.
LLMs operate at scale, produce explicit and reproducible reasoning,
and carry broad implicit knowledge of realistic data distributions
acquired during pre-training.

The core privacy insight is simple.
If $\mathcal{D}$ achieves accuracy close to the chance baseline of
50\%, the synthetic data is \emph{perceptually indistinguishable}
from real data.
This provides empirical evidence of privacy robustness under an
informed adversary.
If $\mathcal{D}$ achieves substantially higher accuracy,
discriminative artifacts are present in the synthetic data.
Such artifacts can be exploited to mount record linkage, membership
inference, or attribute inference attacks.


\subsection{Experimental Conditions}
\label{subsec:conditions}
We define two threat conditions (see Table~\ref{tab:conditions}).
Under \textbf{C1} (table only), the discriminator receives a
Markdown-formatted sample of $N = 20$ rows with column names and
dataset dimensions, modelling an adversary with access to the data
release only.
Under \textbf{C2} (table plus metadata), the same sample is augmented with a structured metadata block computed from the full dataset, including per-column descriptive statistics (mean, standard deviation, quartiles, skewness, kurtosis), Shapiro--Wilk results, top-$k$ category frequencies, Shannon entropy, and a Pearson correlation matrix.
This models a scenario common in statistical agency releases, where summary statistics are published alongside the synthetic data.

\begin{table}[ht]
\centering
\caption{Summary of the two experimental conditions.
Both conditions share the same 20-row table sample.
They differ in the additional context provided to the discriminator.}
\label{tab:conditions}
\small
\begin{tabular}{p{1.2cm}p{4.2cm}p{5.2cm}}
\toprule
\textbf{Cond.} & \textbf{Discriminator input} & \textbf{Threat model} \\
\midrule
C1 & 20-row Markdown table + column names + dataset shape
   & Adversary with access to the data release only \\
\addlinespace
C2 & C1 \textit{plus} per-column statistics (mean, std, quartiles,
    skewness, kurtosis), Shapiro--Wilk results, top-$k$ category
    frequencies, Shannon entropy, Pearson correlations
   & Adversary who also has access to published summary statistics \\
\bottomrule
\end{tabular}
\end{table}

\subsection{Prompt Design}
\label{subsec:prompt}
 
Each trial consists of a system prompt and a user prompt.
The system prompt instructs the LLM to act as a data scientist and
mandates a structured JSON response with five fields: \texttt{verdict}
(\texttt{REAL} or \texttt{SYNTHETIC}), \texttt{confidence} (0--100),
\texttt{reasoning}, \texttt{red\_flags}, and
\texttt{supporting\_evidence}.
Temperature is fixed at zero for reproducibility.
The user prompt instantiates the C1 or C2 template with the 20-row
Markdown table and, for C2, the statistical metadata block.
The structured output allows unambiguous extraction of verdicts and
makes LLM reasoning auditable.
We provide no few-shot examples; discrimination accuracy reflects the
properties of the synthetic data rather than task-specific calibration.

\subsection{Disclosure Risk Score}
\label{subsec:drs}

We introduce the \emph{Disclosure Risk Score}~(DRS) to translate raw discrimination accuracy into a privacy-oriented scalar:

\begin{equation}
  \mathrm{DRS} = \min\!\left(
    \hat{p}_{\mathrm{REAL}},\;
    \hat{p}_{\mathrm{SYNTHETIC}}
  \right),
  \label{eq:drs}
\end{equation}

where $\hat{p}_{\mathrm{REAL}}$ is the fraction of REAL trials correctly labeled \texttt{REAL}, and $\hat{p}_{\mathrm{SYNTHETIC}}$ is the fraction of REAL SYNTHETIC trials correctly labeled \texttt{SYNTHETIC}.
We take the minimum of the two per-class accuracies to account against label-bias artifacts.
A discriminator that labels everything \texttt{REAL} achieves $\hat{p}_{\mathrm{REAL}} = 1$ but $\hat{p}_{\mathrm{SYNTHETIC}} = 0$,
yielding $\mathrm{DRS} = 0$.
This correctly reflects zero discriminative ability.
A DRS close to 0 indicates that the discriminator cannot distinguish real from synthetic data.
A DRS close to 0.5 indicates chance-level balanced discrimination.
A DRS above 0.5 indicates above-chance discrimination, which is the privacy-concerning regime.

\subsection{Human Annotation Protocol}
\label{subsec:human_protocol}
 
We provide a human discrimination baseline using the same C1/C2 trial structure~\footnote{Screen recording: \url{https://github.com/SlokomManel/LLM-as-a-Discriminator/blob/main/results/human/streamlit-human_labeling_app-2026-05-31-01-57-42.webm}}.
Each session presents an annotator with a 20-row Markdown table sampled from a real or synthetic dataset; under C1 the table only is shown, under C2 it is augmented with the metadata block from
Section~\ref{subsec:conditions}.
The annotator records a binary verdict (\texttt{REAL} or
\texttt{SYNTHETIC}), a confidence score (0--100), and optional
free-text notes.
Sessions follow a paired C1 then C2 design: the same 10 samples are
shown first without metadata, then with metadata, enabling
within-session measurement of the information effect without sample
confounding.

\section{Experimental Setup}
\label{sec:setup}
\subsection{Datasets}
We use two census datasets with the same income prediction task but different column encodings.
\textit{UCI Adult census income}~\footnote{ADULT: \url{https://archive.ics.uci.edu/dataset/2/adult}} comprises 30{,}162 records and 14 attributes (continuous, ordinal, and categorical).
The binary sensitive target is income.
Column values are stored as human-readable strings (e.g., occupation = ``Craft-repair''), making the data directly interpretable by an LLM.
\textit{ACS Census Income 2018} comprises 32{,}561 records and 10 attributes.
We use the income-above-\$50K prediction task.
Unlike UCI Adult, all categorical columns are stored as numeric integer codes.
This presents a different perceptual challenge to the LLM discriminator.

\subsection{Synthetic Data Generation}
 
We generate synthetic data using three methods via the SDV
framework~\footnote{SDV: \url{https://docs.sdv.dev/sdv}} with default
hyperparameters: CTGAN~\cite{NIPS2019ctgan}, a conditional GAN with
mode-specific normalization; TVAE~\cite{NIPS2019ctgan}, a variational
autoencoder over the joint distribution; and Gaussian
Copula~\cite{patki2016synthetic}, a parametric baseline using
marginal-specific transformations.
Each method produces a synthetic dataset of the same dimensions as
the real data.

\subsection{LLM models evaluated}
We evaluate two LLM families via cloud APIs.
\textbf{Gemini}: We use \texttt{gemini-2.5-flash}~\footnote{Gemini: \url{https://ai.google.dev/gemini-api/docs/models/gemini-2.5-flash}}, a lightweight multimodal model developed by Google that supports structured JSON output and long-context reasoning.
\textbf{LLaMA (via Groq)}: We use \texttt{llama-3.1-8b-instant}~\footnote{LLaMA: \url{https://console.groq.com/docs/model/llama-3.1-8b-instant}}, an open-weight model developed by Meta and served via the Groq inference API.

\subsection{Trial design}
\label{subsec:trial}

For each synthesizer and condition pair we planned $T = 200$ trials,
stratified equally between real and synthetic tables.
Each trial samples 20 rows uniformly at random.
Trial order and labels are randomized to prevent position bias.
Each trial yields one structured JSON response containing a binary verdict, a confidence score, and free-text reasoning.

\textit{API failures and final sample size.}
For UCI Adult, we planned 3{,}600 trials in total
(3 synthesizers $\times$ 2 conditions $\times$ 200 trials per cell).
Of these, \textbf{3{,}263 calls (90.6\%) failed} due to HTTP 429 quota
errors and model deprecations.
The final UCI Adult sample comprises \textbf{337 valid verdicts}:
287 from \texttt{llama-3.1-8b-instant} and 50 from
\texttt{gemini-2.5-flash}.
For ACS Census, we used the same two models with $T = 10$ trials per
cell, yielding \textbf{114 valid verdicts} (54 from Gemini, 60 from
Groq).
Table~\ref{tab:trial_counts} reports the trial counts per cell.

\begin{table}[t]
\centering
\caption{Trial counts per synthesizer and condition cell after
discarding API failures.}
\label{tab:trial_counts}
\small
\begin{tabular}{lcccc}
\toprule
 & \multicolumn{2}{c}{\textbf{UCI Adult}}
 & \multicolumn{2}{c}{\textbf{ACS Census}} \\
\cmidrule(lr){2-3}\cmidrule(lr){4-5}
\textbf{Method} & \textbf{C1} & \textbf{C2}
                & \textbf{C1} & \textbf{C2} \\
\midrule
CTGAN           & 62 & 62 & 19 & 20 \\
TVAE            & 57 & 52 & 19 & 19 \\
Gaussian Copula & 54 & 50 & 18 & 19 \\
\midrule
\textit{Total}  & 173 & 164 & 56 & 58 \\
\bottomrule
\end{tabular}
\end{table}

\subsection{Formal privacy baselines}
\label{subsec:baselines}

We validate the LLM-based DRS against two empirical privacy measures.
\textit{Classifier two-sample test (C2ST):}
We train a logistic regression classifier on a balanced pool of 70\% of real and synthetic records and evaluate its discrimination accuracy on the held-out 30\%.
The C2ST score is the overall accuracy of this classifier, measuring whether the two populations are statistically separable.
This is a distributional detection test, not a membership inference attack: it asks whether real and synthetic records come from the same distribution, not whether any individual record was in the generator's training set.
\textit{Record linkage risk:}
For each synthetic record, we compute the Euclidean distance to its
nearest real record in feature space after min-max normalization.
The linkage success rate is the fraction of synthetic records whose
nearest-neighbor distance falls within the 10th percentile threshold.


\subsection{Human annotation study}
\label{subsec:human_setup}
Two human annotators (HA1 and HA2) independently completed six annotation sessions each: 2 datasets $\times$ 3 synthesis methods,
covering both C1 and C2 conditions with 10 trials per session
(5 real and 5 synthetic).
The study yields \textbf{240 unique trials} (120 per annotator).
Both annotators achieved full coverage across methods, conditions, and
datasets.
We apply the DRS formula (Equation~\ref{eq:drs}) identically to human
and LLM verdicts, enabling direct comparison.

\section{Experimental Results}
\label{sec:results}

We report results in five steps: baseline LLM discrimination,
metadata effects, method ranking, alignment with formal baselines,
and calibration against human judgment.
All DRS values are reported per provider.

\subsection{RQ1: Baseline LLM Discrimination}
Tables~\ref{tab:drs_main} and~\ref{tab:census_drs} report per-provider DRS for both datasets by synthesis method and condition, under the minimum-coverage matched protocol ($N \geq 8$ per cell,
$N_\text{Groq} = N_\text{Gemini}$ within each included cell).
The two providers show contrasting and dataset-dependent behavior.


\begin{table}[t]
\centering
\caption{UCI Adult DRS by provider, synthesis method, and condition
(minimum-coverage matched protocol, $N_\text{Groq}=N_\text{Gemini}$ per cell).
Groq labels every table as \textsc{Real} ($\hat{p}_\text{REAL}=100\%$,
$\hat{p}_\text{SYN}=0\%$), collapsing DRS to~0\%.
Gemini discriminates perfectly for CTGAN and TVAE under both conditions.
Gaussian Copula cells excluded ($N_\text{Gemini}<8$).}
\label{tab:drs_main}
\small
\begin{tabular}{lllcccc}
\toprule
\textbf{Provider} & \textbf{Method} & \textbf{Cond.} & \textbf{$N$} &
  \textbf{$\hat{p}_\text{REAL}$ (\%)} &
  \textbf{$\hat{p}_\text{SYN}$ (\%)} &
  \textbf{DRS (\%)} \\
\midrule
\multirow{6}{*}{Groq}
  & CTGAN           & C1 & 12 & 100.0 &  0.0 &  0.0 \\
  & CTGAN           & C2 & 12 & 100.0 &  0.0 &  0.0 \\
  & TVAE            & C1 & 11 & 100.0 &  0.0 &  0.0 \\
  & TVAE            & C2 & 11 & 100.0 &  0.0 &  0.0 \\
  & Gaussian Copula & C1 & -- & --    & --   & --   \\
  & Gaussian Copula & C2 & -- & --    & --   & --   \\
\midrule
\multirow{4}{*}{Gemini}
  & CTGAN           & C1 & 12 & 100.0 & 100.0 & \textbf{100.0} \\
  & CTGAN           & C2 & 12 & 100.0 & 100.0 & \textbf{100.0} \\
  & TVAE            & C1 & 11 & 100.0 & 100.0 & \textbf{100.0} \\
  & TVAE            & C2 & 11 & 100.0 & 100.0 & \textbf{100.0} \\
\bottomrule
\end{tabular}
\end{table}

\begin{table}[t]
\centering
\caption{ACS Census DRS by provider, synthesis method, and condition
(minimum-coverage matched protocol, $N_\text{Groq}=N_\text{Gemini}$ per cell).
Groq labels every table as \textsc{Synthetic} ($\hat{p}_\text{REAL}=0\%$,
$\hat{p}_\text{SYN}=100\%$), the inverse collapse from Adult.
Gemini achieves DRS$=100\%$ under C1 for all three methods;
under C2 CTGAN drops to DRS$=0\%$ and TVAE to DRS$=25\%$.}
\label{tab:census_drs}
\small
\begin{tabular}{lllcccc}
\toprule
\textbf{Provider} & \textbf{Method} & \textbf{Cond.} & \textbf{$N$} &
  \textbf{$\hat{p}_\text{REAL}$ (\%)} &
  \textbf{$\hat{p}_\text{SYN}$ (\%)} &
  \textbf{DRS (\%)} \\
\midrule
\multirow{6}{*}{Groq}
  & CTGAN           & C1 & 10 &   0.0 & 100.0 &   0.0 \\
  & CTGAN           & C2 & 10 &   0.0 & 100.0 &   0.0 \\
  & TVAE            & C1 & 10 &   0.0 & 100.0 &   0.0 \\
  & TVAE            & C2 & 10 &   0.0 & 100.0 &   0.0 \\
  & Gaussian Copula & C1 & 10 &   0.0 & 100.0 &   0.0 \\
  & Gaussian Copula & C2 & 10 &   0.0 & 100.0 &   0.0 \\
\midrule
\multirow{6}{*}{Gemini}
  & CTGAN           & C1 &  9 & 100.0 & 100.0 & \textbf{100.0} \\
  & CTGAN           & C2 & 10 & 100.0 &   0.0 &   0.0 \\
  & TVAE            & C1 &  9 & 100.0 & 100.0 & \textbf{100.0} \\
  & TVAE            & C2 &  9 & 100.0 &  25.0 &  25.0 \\
  & Gaussian Copula & C1 &  8 & 100.0 & 100.0 & \textbf{100.0} \\
  & Gaussian Copula & C2 &  9 & 100.0 & 100.0 & \textbf{100.0} \\
\bottomrule
\end{tabular}
\end{table}

\textit{}{Groq (LLaMA-3.1-8b).}
Groq yields DRS = 0\% across all reported cells on both datasets via two opposing mechanisms.
On UCI Adult, it labels 98.3\% of tables as \texttt{REAL}, defaulting to the majority label when no clear artifact is present.
On ACS Census, the bias inverts: every table is labeled
\texttt{SYNTHETIC} because genuine ACS properties such as heavy-tailed income and many unique categorical values are flagged as synthetic artifacts, even though these appear in both real and synthetic records.
In both cases DRS = 0\% reflects a collapsed discriminator, not good privacy protection.
 
\textit{Google (Gemini-2.5-Flash).}
On UCI Adult, Gemini achieves DRS = 100\% for CTGAN and TVAE under
both conditions.
On ACS Census, DRS = 100\% under C1 for all three synthesizers but
degrades under C2: CTGAN drops to 0\%, TVAE to 25\%, while Gaussian
Copula holds at 100\%.
This suggests that distributional metadata disrupts the detection
signal when real and synthetic marginals are closely aligned.
 
\textit{Overconfidence.}
Mean self-reported LLM confidence is 82--99\% regardless of verdict
correctness~\cite{chhikaramind}.
Raw confidence scores should not be used as a privacy signal without calibration.

\subsection{RQ2 and RQ3: Metadata effects and method ranking}
Because LLaMA (via Groq) collapses to DRS = 0\% under all conditions and methods, neither the C1 to C2 metadata effect nor the cross-method ranking is estimable from its outputs.
We therefore focus on Gemini.

On UCI Adult, metadata has no effect on Gemini: DRS = 100\% under both C1 and C2 for CTGAN and TVAE.
On ACS Census, metadata reduces discrimination for CTGAN (100\% to 0\%) and TVAE (100\% to 25\%), while Gaussian Copula is unaffected.
This shows that the effect of distributional metadata is
dataset- and method-dependent, not uniformly risk-increasing.

For method ranking, Gemini achieves DRS = 100\% in most cells, with
condition-sensitive reversals on Census C2 for CTGAN and TVAE.
No synthesis method dominates across all providers and conditions
combinations.
Privacy rankings are only meaningful when reported per provider and
per condition.

\subsection{RQ4: Alignment with formal privacy measures}
Table~\ref{tab:correlation} compares Gemini DRS against two formal
privacy baselines across both datasets and all synthesis methods.
We use Gemini only, as Groq provides no meaningful discrimination signal.
We observe a ceiling effect on the UCI Adult dataset.
Gemini achieves DRS = 100\% for CTGAN and TVAE under both conditions.
This ceiling prevents any DRS-based method ranking.
C2ST still differentiates: TVAE (0.725) is well above chance, while
CTGAN (0.538) and Gaussian Copula (0.510) are near chance.
Yet all three are equally detectable by Gemini.
As for ACS census, we see a partial alignment. 
Under C1, all methods reach DRS = 100\%.
Under C2, DRS degrades for CTGAN (100\% to 0\%) and TVAE
(100\% to 25\%), while Gaussian Copula holds at 100\%.
Within C2, the DRS ordering (Gaussian Copula $>$ TVAE $>$ CTGAN)
is inverted relative to C2ST (TVAE $>$ CTGAN $>$ Gaussian Copula).
Gaussian Copula has a low C2ST score (0.542) yet consistently high DRS, suggesting its distributional artifacts remain perceptually salient regardless of metadata context.

 \begin{table}[t]
\centering
\caption{Gemini DRS per condition versus formal privacy baselines on UCI Adult and ACS Census.
Gaussian Copula on UCI Adult is excluded ($N_\text{Gemini}<8$).
C2ST = accuracy of a logistic regression separating real from synthetic records.
Linkage rate = fraction of synthetic records within the 10th-percentile nearest-neighbor distance to real data.}
\label{tab:correlation}
\small
\begin{tabular}{llcccc}
\toprule
\textbf{Dataset} & \textbf{Method} &
  \textbf{DRS C1 (\%)} & \textbf{DRS C2 (\%)} &
  \textbf{C2ST} & \textbf{Linkage (\%)} \\
\midrule
\multirow{3}{*}{UCI Adult}
  & CTGAN           & 100.0 & 100.0 & 0.538 & 87.2 \\
  & TVAE            & 100.0 & 100.0 & 0.725 & 87.8 \\
  & Gaussian Copula & --    & --    & 0.510 & 91.6 \\
\midrule
\multirow{3}{*}{ACS Census}
  & CTGAN           & 100.0 &   0.0 & 0.566 & 85.1 \\
  & TVAE            & 100.0 &  25.0 & 0.659 & 88.4 \\
  & Gaussian Copula & 100.0 & 100.0 & 0.542 & 84.3 \\
\bottomrule
\end{tabular}
\end{table}
 

\subsection{RQ5: Human annotation baseline}
\label{subsec:human_results}
 
\begin{figure}[t]
\centering
\includegraphics[width=0.7\textwidth]{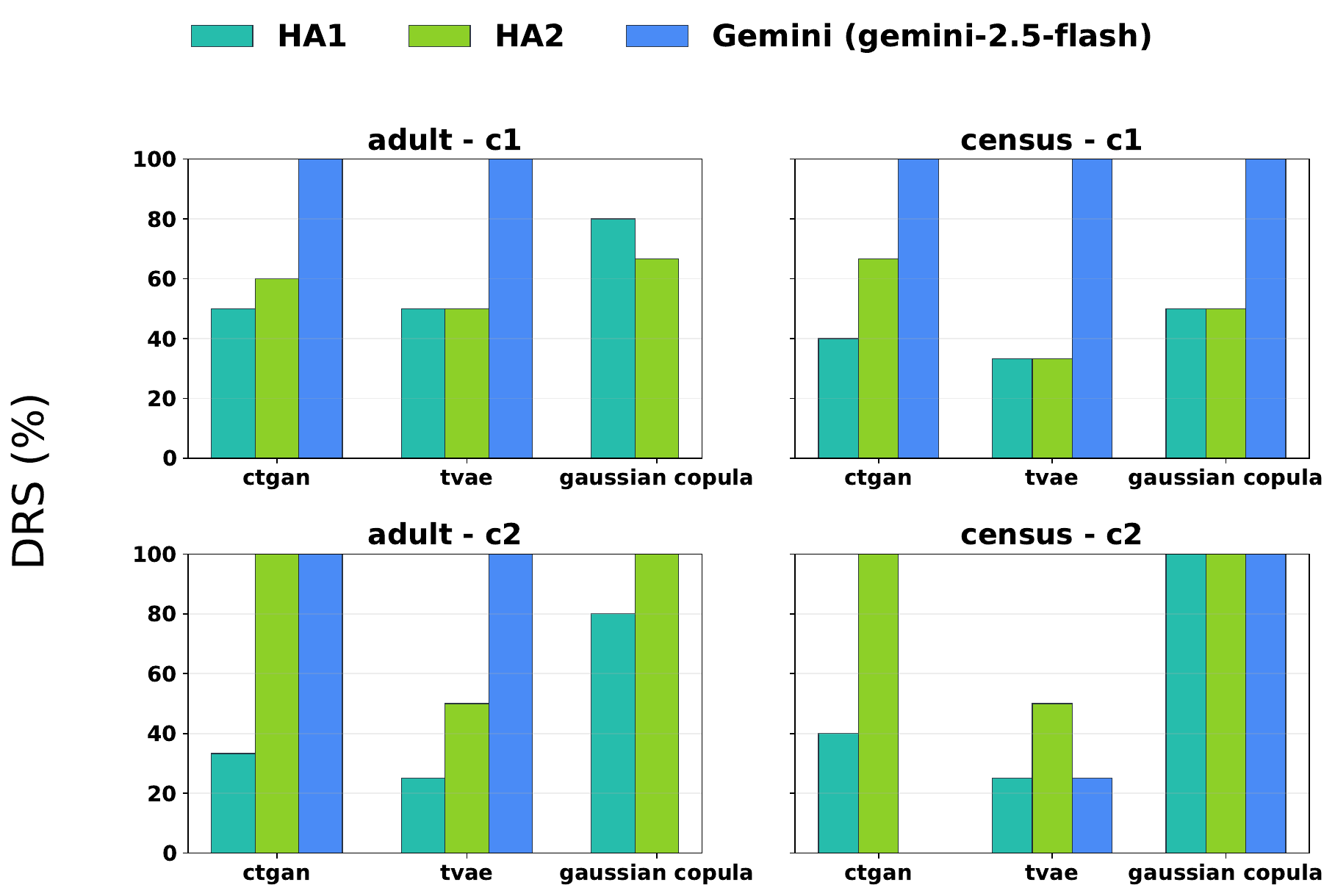}
\caption{Cell-level DRS by dataset, synthesis method, and condition
(C1/C2) for HA1, HA2, and Gemini (\texttt{gemini-2.5-flash}).}
\label{fig:human_vs_llm}
\end{figure}

Figure~\ref{fig:human_vs_llm} reports the human and LLM comparison
with explicit model labels, split by condition (C1 and C2), and broken down by dataset and synthesis method.
 
Multiple findings emerge.
First, human annotators outperform LLaMA.
Both annotators substantially exceed LLaMA-3.1-8b across both datasets (pooled DRS = 66.4\% vs.\ LLaMA $\approx$ 0\%).
On UCI Adult, LLaMA achieves DRS = 0\% in five of six cells due to
REAL-label bias.
On ACS Census it achieves DRS = 0\% in all six cells due to
SYNTHETIC-label bias.
Human annotators correctly identify synthetic records in both settings.
Second, Gemini matches or exceeds human performance.
For CTGAN on UCI Adult, Gemini achieves DRS = 100\% under both
conditions, above both annotators (HA1: 50\% and 33\%, HA2: 60\%
and 100\% under C1 and C2 respectively).
On ACS Census, Gemini achieves DRS = 100\% for all three synthesizers under C1, exceeding human performance in most cells.
Frontier LLMs can match or surpass trained human judgment on this task.
Next, the inter-annotator agreement is moderate.
The mean absolute DRS difference between annotators is 20.6 pp.
HA2 achieves consistently higher DRS (77.9\%) and accuracy (80.0\%).
Both annotators show above-chance confidence calibration: correct
predictions carry a higher mean confidence than incorrect ones
(HA1: 62.3\% vs.\ 53.6\%; HA2: 58.9\% vs.\ 50.5\%).
Finally, both annotators independently surface the same structural artifacts
as Gemini: broken \texttt{education}/\texttt{education\_num} mappings,
gender and relationship contradictions, and implausible age and
education combinations.
We confirm that structural constraint violations are the primary
perceptual discriminator for both humans and frontier LLMs.

\section{Discussion}
\label{sec:discussion}

Our main takeaway is practical: LLM discrimination is a useful
first-pass privacy screen only when interpreted per model.
In several cells, discriminator capability dominates synthesis-method effects.
We use DRS as a screening metric, not as a standalone privacy guarantee.

\textbf{Connecting DRS to traditional privacy measures}
Table~\ref{tab:comparison} situates LLM-based DRS in the broader
privacy evaluation landscape.
Each metric targets a different privacy dimension: DP gives worst-case guarantees but requires white-box access;
k-anonymity checks structural grouping but misses artifacts outside quasi-identifiers;
MIA~\cite{Shokri2017Membership} targets record-level memorization but requires shadow-model training; record linkage measures feature-space proximity but not semantic plausibility.
DRS captures perceptual discriminability without model access and
produces interpretable reasoning.
No single metric is sufficient.
We recommend DRS and C2ST as fast distributional screens, followed by MIA when record-level memorization is a concern.

\begin{table}[ht]
\centering
\caption{Comparison of privacy evaluation paradigms.}
\label{tab:comparison}
\small
\begin{tabular}{lcccc}
\toprule
\textbf{Measure} & \textbf{Model access} & \textbf{Compute} &
  \textbf{Interpretable} & \textbf{Black-box} \\
\midrule
Differential Privacy $\varepsilon$ & Required & Low      & Formal   & No      \\
$k$-Anonymity                      & Not req. & Low      & Yes      & Yes     \\
MIA success rate                   & Optional & High     & Moderate & Partial \\
Record Linkage Risk                & Not req. & Moderate & Yes      & Yes     \\
\textbf{LLM-based DRS (ours)}      & Not req. & Low      & Yes      & Yes     \\
\bottomrule
\end{tabular}
\end{table}


\textbf{LLM Reasoning Patterns}
\label{sec:reasoning}
We qualitatively analyzed all 337 valid UCI Adult verdicts.
Gemini cites one decisive structural violation per verdict.
LLaMA lists generic statistical flags that appear in both real and
synthetic records, which explains its REAL-label bias.

\paragraph{Structural constraint violations.}
Gemini's primary \texttt{SYNTHETIC} cue is the broken dependency
between \texttt{education} and \texttt{education\_num} (e.g.,
``Bachelors'' $\to$ 13, ``10th'' $\to$ 6 in UCI Adult).
CTGAN and TVAE break this mapping; Gemini flags it as decisive:
\textit{```education\_num' does not maintain a 1-to-1 mapping with
`education'; e.g., `10th' grade is assigned 4, not 6.''}.
We also observe a marital status and relationship contradiction:
CTGAN pairs ``Never-married'' with ``Husband'' in 4.5\% of records.

\paragraph{Capital gain/loss anomalies.}
UCI Adult has highly sparse capital gain and loss fields (over 90\%
zeros, ceiling of 99{,}999).
Synthetic generators fill in small non-zero values; LLaMA flags:
\textit{``High skewness in `capital\_gain' and `capital\_loss';
large number of zeros in capital\_gain.''}.

\section{Conclusion, Limitations, and Future Work}
\label{sec:conclusion}
 
We proposed and evaluated an LLM-as-Discriminator protocol for
privacy auditing of synthetic tabular data (451 valid trials, two
datasets, three synthesis methods, two model families).
The core finding is simple: the discriminator model matters as much
as the synthesizer.
On UCI Adult, Gemini achieves DRS = 100\% for CTGAN and TVAE while
LLaMA collapses to DRS = 0\% via REAL-label bias.
On ACS Census, the same split appears in reverse.
Pooled cross-model DRS is misleading and should not be the primary
reported result.
DRS shows directional consistency with formal baselines, and the
human pilot calibrates the LLM signal: pooled human DRS is 66.4\%,
well above LLaMA, with Gemini matching or exceeding human performance.
Model capability is the bottleneck, not the LLM-as-discriminator
paradigm itself.
 
Our study has four main limitations.
First, 90.6\% of planned trials failed due to API quota exhaustion,
leaving per-cell $N$ as low as 9; future work should use locally
hosted models.
Second, most valid verdicts come from a single 8B-parameter model;
more capable models may behave differently.
Third, DRS is measured under a single zero-shot prompt; chain-of-thought
or few-shot designs remain unexplored.
Fourth, we cover three synthesis methods on two datasets;
diffusion-based~\cite{kotelnikov2023tabddpm} and LLM-based
synthesizers~\cite{borisov2023GReat} remain unevaluated.
 
The most important next steps are:
(i) encoding-aware auditing via automatic column decoding;
(ii) broader validation across more datasets and synthesis methods;
(iii) stable local model deployments to avoid API sparsity;
(iv) larger human annotation studies; and
(v) joint privacy-utility reporting.

\appendix
\section{Appendix}
\label{sec:appendix}
\subsection{Matched-Sampling Robustness (Seed Sweep)}
\label{app:seed}

\begin{figure}[h]
\centering
\includegraphics[width=0.95\textwidth]{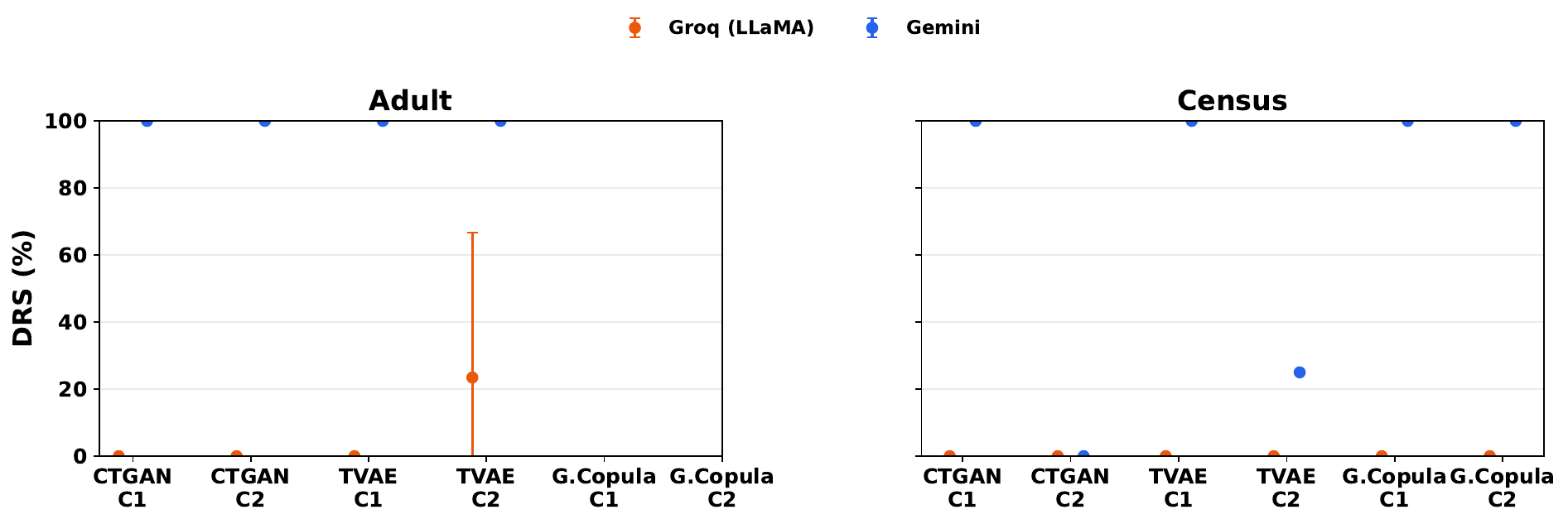}
\caption{Cell-level DRS stability over 50 random seeds.
Points show mean DRS; whiskers show empirical 95\% intervals.
Each point is annotated with matched per-cell sample size $N$.
Uncertainty is concentrated in Adult TVAE C2 for Groq.}
\label{fig:seed_stability}
\end{figure}

We swept 50 independent matched-downsampling seeds (1000--1049) and
recomputed per-cell DRS across all included cells.
Figure~\ref{fig:seed_stability} reports empirical 95\% intervals.
Results are stable: 19 of 20 cells have effectively zero interval
width.
The single exception is Adult TVAE C2 under Groq
(mean DRS = 22.1\%, interval [0.0, 73.1] pp, $N = 11$).
Mean interval width is 0.0 pp for Gemini and 7.3 pp for Groq.
The core finding (Gemini: high discriminability; Groq: collapse) is
seed-independent.
Low-$N$ Groq cells warrant uncertainty-aware interpretation.

\subsection{Per-Model Accuracy and Label Bias}
\label{app:bias}

\begin{figure}[h]
\centering
\includegraphics[width=0.9\textwidth]{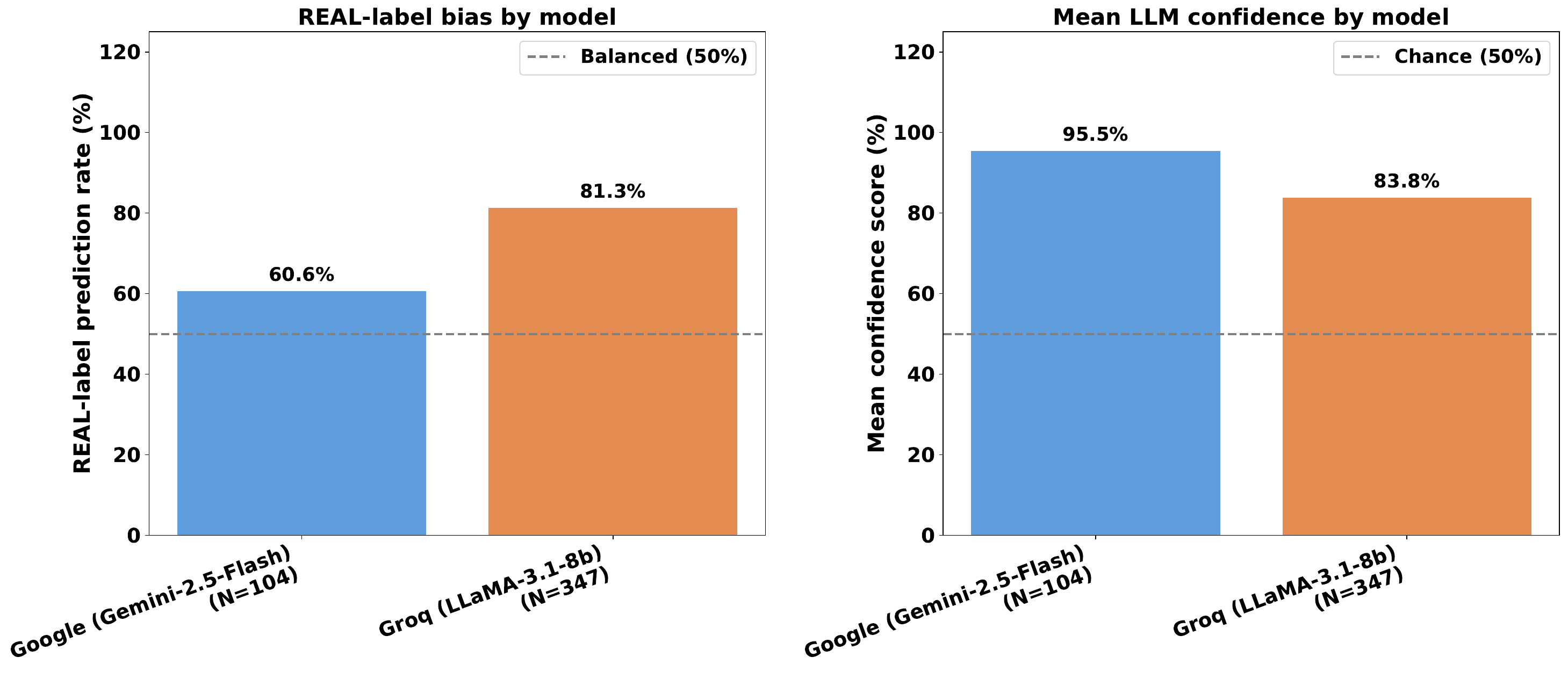}
\caption{REAL-label prediction rate (left) and mean LLM confidence
(right) by model.
LLaMA labels 98.3\% of Adult tables as \texttt{REAL};
Gemini shows lower label bias and higher accuracy.}
\label{fig:bias_confidence}
\end{figure}

Table~\ref{tab:model_accuracy} reports overall accuracy and REAL-label
prediction rate per model on UCI Adult.
Figure~\ref{fig:bias_confidence} shows REAL-label prediction rate and
mean confidence by model.

\begin{table}[h]
\centering
\caption{Per-model discrimination accuracy and REAL-label prediction
rate on UCI Adult (337 valid verdicts).}
\label{tab:model_accuracy}
\small
\begin{tabular}{llccc}
\toprule
\textbf{Family} & \textbf{Model} & \textbf{$N$} &
  \textbf{Acc.\ (\%)} & \textbf{REAL-pred (\%)} \\
\midrule
Open-weight   & LLaMA-3.1-8b-instant & 287 & 51.9  & 98.3 \\
Google Gemini & Gemini-2.5-Flash      &  50 & 100.0 & 56.0 \\
\bottomrule
\end{tabular}
\end{table}

\subsection{Reasoning Theme Prevalence}
\label{app:reasoning}

Table~\ref{tab:reasoning_themes} reports the eight dominant reasoning
themes identified in the qualitative analysis of 337 UCI Adult verdicts,
with prevalence broken down by provider.

\begin{table}[h]
\centering
\caption{Reasoning theme prevalence by provider (UCI Adult,
337 valid verdicts). Percentages show the fraction of each provider's
records in which the theme appears in \texttt{red\_flags} or
\texttt{reasoning} text.}
\label{tab:reasoning_themes}
\small
\begin{tabular}{lrr}
\toprule
\textbf{Reasoning theme} & \textbf{Gemini (\%)} & \textbf{LLaMA (\%)} \\
\midrule
Capital gain/loss distribution anomaly & 1.8 & 23.9 \\
General distributional irregularities  & 1.8 & 23.9 \\
Syntactic/structural artifact          & 1.8 & 22.7 \\
Education/education\_num mapping       & 2.0 & 15.7 \\
Dataset fingerprinting (UCI Adult)     & 1.8 & 12.2 \\
Marital/relationship contradiction     & 1.4 & 10.7 \\
Age distribution anomaly               & 0.0 & 10.2 \\
Zero/sparse capital values             & 1.5 &  2.1 \\
\bottomrule
\end{tabular}
\end{table}

%
%
\bibliographystyle{splncs04}
\bibliography{mybibliography}

\end{document}